\documentclass{article}

\usepackage[preprint]{neurips_2023}

\usepackage[utf8]{inputenc} 
\usepackage[T1]{fontenc}    
\usepackage{hyperref}       
\usepackage{url}            
\usepackage{booktabs}       
\usepackage{amsfonts}       
\usepackage{nicefrac}       
\usepackage{microtype}      
\usepackage{xcolor}         
\usepackage{diagbox}
\usepackage{todonotes}
\usepackage{colortbl}
\usepackage{array}
\usepackage{amsmath}
\usepackage{multirow}
\usepackage{makecell}
\usepackage{multicol}
\usepackage{graphicx}
\usepackage{algorithmicx,algorithm}
\usepackage[noend]{algpseudocode}
\usepackage{color}
\usepackage{svg}

\usepackage{caption}
\usepackage{wrapfig}

\title{Is Synthetic Data From Diffusion Models Ready for Knowledge Distillation?}

\author{
    Zheng Li \textsuperscript{\rm 1}\thanks{Equal contributions.} ,
    Yuxuan Li \textsuperscript{\rm 1}\footnotemark[1] ,
    Penghai Zhao \textsuperscript{\rm 1},
    Renjie Song \textsuperscript{\rm 2},
    Xiang Li \textsuperscript{\rm 1}\thanks{Corresponding author.} ,
    Jian Yang \textsuperscript{\rm 1}\footnotemark[2]\\
    \\
    \textsuperscript{\rm 1} Nankai University,
    \textsuperscript{\rm 2} Megvii Technology \\
    \\
    \texttt{\{zhengli97, zhaopenghai\}@mail.nankai.edu.cn, yuxuan.li.17@ucl.ac.uk} \\ \texttt{songrenjie@megvii.com, \{xiang.li.implus, csjyang\}@nankai.edu.cn}
}

\begin{document}

\maketitle
\begin{abstract}

Diffusion models have recently achieved astonishing performance in generating high-fidelity photo-realistic images. Given their huge success, it is still unclear whether synthetic images are applicable for knowledge distillation when real images are unavailable. In this paper, we extensively study whether and how synthetic images produced from state-of-the-art diffusion models can be used for knowledge distillation without access to real images, and obtain three key conclusions: (1) synthetic data from diffusion models can easily lead to state-of-the-art performance among existing synthesis-based distillation methods, (2) low-fidelity synthetic images are better teaching materials, and (3) relatively weak classifiers are better teachers. Code is available at \url{https://github.com/zhengli97/DM-KD}.

\end{abstract}

\section{Introduction}
\label{section:intro}




Knowledge distillation~\cite{hinton2015distilling} aims to train a lightweight student model on a target dataset under the supervision of a pre-trained teacher model. Various forms~\cite{romero2014fitnets,zagoruyko2016paying,kim2018paraphrasing,chen2021distilling,chen2022improved,li2022curriculum} and paradigms~\cite{zhang2018deep,zhu2018knowledge,zhang2019your,chen2020online,kim2021self,qian2022switchable} have been proposed to improve the efficiency of distillation. 
However,  training datasets might not always be available due to security and privacy concerns, which makes existing data-dependent distillation methods no longer applicable. 
To address this issue, several synthesis-based distillation methods are proposed. These methods utilize either white-box teacher statistics~\cite{lopes2017data,chen2019data,yin2020dreaming,fang2022up} or data augmentation techniques~\cite{asano2023the} to generate synthetic samples that serve as proxy training datasets for distillation. By training on such synthetic data, the student model can successfully learn from the teacher model without access to real training data.

Recently, diffusion models~\cite{ho2020denoising,saharia2022imagen,ramesh2021zero,ramesh2022hierarchical} are attracting increasing attention in image generation tasks. Several high-performance diffusion models, including GLIDE~\cite{nichol2021glide}, Stable Diffusion~\cite{rombach2022high}, and DiT~\cite{peebles2022scalable}, have demonstrated impressive abilities to generate high-fidelity photo-realistic images at high resolutions. This leads to a natural question: Can these synthetic images be used for downstream tasks? 
He et al.~\cite{he2023is} made the first attempt using synthetic data to improve the zero-shot and few-shot recognition performance of a classifier. Azizi et al.~\cite{azizi2023synthetic} utilized the diffusion model as a generative data augmentation method to increase the scale of existing datasets.
However, to our best knowledge, few works have explored the impact of diffusion generative models on knowledge distillation.


In this paper, we aim to investigate whether and how synthetic images generated from state-of-the-art diffusion models can be utilized for knowledge distillation without access to real images. Through our research, we have identified three key findings, which are outlined below:

\textbf{(1) Synthetic data from diffusion models can easily lead to state-of-the-art performance among existing synthesis-based distillation methods.}
Our proposed approach differs from previous synthesis-based distillation methods by eliminating the need to design complex and precise generative methods based on the white-box teacher model or the careful selection of a single datum. Instead, we leverage publicly available state-of-the-art diffusion models to generate a proxy dataset for distillation, even when the teacher model is a black box.
As a result, our method achieves state-of-the-art performance across multiple datasets, including ImageNet-1K, ImageNet-100, CIFAR-100, and Flowers-102 datasets. Notably, our method can effectively transfer knowledge in the target dataset from the teacher model to the student model, even when there is no intersection of categories between the synthetic dataset and the target dataset.


\textbf{(2) Low-fidelity synthetic images are better teaching materials.}
While existing diffusion models have demonstrated an impressive ability to generate high-fidelity and realistic images, our experiments suggest that higher image fidelity does not necessarily lead to better distillation performance. On the contrary, our experimental results indicate that relatively low-fidelity images, as assessed by the ImageReward score~\cite{xu2023imagereward}, are better learning materials for students during the distillation process. 
Specifically, for conditional diffusion models, our experiments suggest that the relatively \emph{small} guidance scaling factor~($s$) and sampling time step~($T$) are better synthetic dataset generators for knowledge distillation, despite the fact that larger $s$ and $T$ can result in higher ImageReward scores and visually appealing/appropriate images.

\textbf{(3) Relatively weak classifiers are better teachers.} On real datasets, knowledge distillation is typically achieved by distilling a larger teacher model~(e.g., ResNet34) to a smaller student model~(e.g., ResNet18). However, on synthetic datasets, we observe the opposite phenomenon, where relatively weak classifiers tend to perform better than strong classifiers.
Specifically, when training ResNet34 on the synthetic dataset, using ResNet18 as the teacher model leads to a 3\% improvement in performance compared to using ResNet50 as the teacher model. 

\section{Related Work}

\textbf{Conditional Diffusion Models.}
Diffusion model~\cite{sohl2015deep,ho2020denoising,nichol2021improved} is a type of generative model that learns to model the probability distribution of a dataset by gradually adding sampled Gaussian noise to the data to "diffuse" the data and then trying to recover the original data by denoising the noise step by step until high-quality images are obtained. Conditional diffusion models is a type of diffusion model that is conditioned on some additional information, such as a text prompt~\cite{saharia2022imagen,nichol2021glide,rombach2022high,ramesh2022hierarchical} or a class label~\cite{ho2022cascaded,peebles2022scalable}. The conditioning signal is used to guide the denoising process so that the model can generate samples conditioned on a specific input. 
Recent text-to-image generation models based on diffusion, such as Imagen~\cite{saharia2022imagen}, GLIDE~\cite{nichol2021glide}, SD~\cite{rombach2022high}, and class label conditional model DiT~\cite{peebles2022scalable}, have demonstrated the ability to generate highly realistic and visually stunning images, highlighting the potential power of diffusion models in generative modeling. Its outstanding performance makes it a popular choice for a variety of low-level image processing tasks, such as super-resolution~\cite{li2022srdiff,saharia2022image}, image inpainting~\cite{song2020score,lugmayr2022repaint,alt2022learning}, and dehazing~\cite{ozdenizci2023restoring}. However, the extent to which generative diffusion models can contribute to high-level tasks has yet to be well explored. 
Several recent studies utilize well-trained open vocabulary text-to-image diffusion models as synthetic data generators for classification tasks. For example, He et al.~\cite{he2022synthetic} show that synthetic data generated by diffusion models can improve model pre-training, as well as zero-shot and few-shot image classification performance. Trabucco et al.~\cite{trabucco2023effective} augment images by editing them using a pre-trained text-to-image diffusion model to bring more semantic attribute diversity, leading to improvements in few-shot settings. Additionally, Azizi et al.~\cite{azizi2023synthetic} demonstrate that finetuned class-label conditional diffusion models can produce high-quality, in-distribution synthetic data that, when trained alongside real data, can achieve state-of-the-art classification performance. 
However, to our best knowledge, few works have explored the impact of diffusion generation models on knowledge distillation.

\textbf{Synthesis-based Knowledge Distillation.}
In recent years, knowledge distillation~\cite{hinton2015distilling} has received increasing attention from the research community, and it has been widely utilized in various vision tasks~\cite{liu2019structured,wang2020intra,zhang2019fast,li2021online,chen2022improved,yang2022vitkd}.
Conventional distillation methods~\cite{park2019relational,chen2020online,li2020online,yang2021knowledge,zhao2022decoupled,li2022curriculum} require labeled training sets to train the student model by transferring knowledge from a large pre-trained teacher model. However, the original training dataset might not always be available due to privacy and safety concerns, making existing data-dependent methods hard to apply to data-scarce scenarios. Synthetic-based distillation methods~\cite{lopes2017data,chen2019data,yin2020dreaming,luo2020large,haroush2020knowledge,binici2022robust,fang2022up,asano2023the} are proposed to solve the data-dependent problem. It typically follows a distilling-by-generating paradigm~\cite{fang2022up} wherein a proxy dataset will be synthesized by utilizing different generative methods. Lopes et al.~\cite{lopes2017data} first proposes to reconstruct the dataset from the metadata~(e.g., activation statistics). DeepInversion~\cite{yin2020dreaming} further optimizes the metadata-based synthesis method by introducing a feature regularization term. FastDFKD~\cite{fang2022up} proposes to speed up the generation process by using common features.
In contrast to previous GAN- and inversion-based distillation methods, One-Image-Distill~(OID)~\cite{asano2023the} utilizes data augmentation to construct a proxy dataset based on a single real image.
Our method stands out from previous approaches that rely on complex generative processes that are based on white-box teachers or require careful selection of the individual real image. These methods are often time-consuming and require a significant amount of effort. In contrast, our approach is simpler, more efficient, and more effective. It only requires the use of publicly available diffusion models for data synthesis.

\section{Method}
\begin{figure*}[t]
	\centering
	\includegraphics[width=1\linewidth]{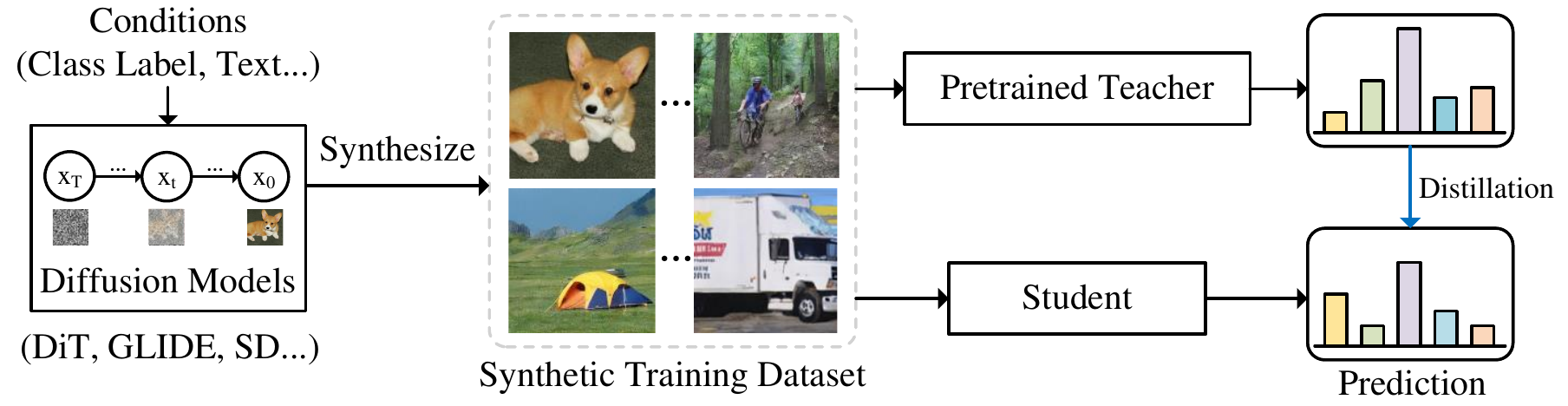}
	\caption{An overview of our proposed synthetic data Knowledge Distillation approach based on Diffusion Models~(DM-KD). We propose to use the diffusion model to synthesize images given the target label space when the real dataset is not available. The student model is optimized to minimize the prediction discrepancy between itself and the teacher model on the synthetic dataset.}
	\label{fig:framework}
        \vspace{-12pt}
\end{figure*}

\textbf{Data Generation from Diffusion Model.}
In recent years, diffusion models have been demonstrated to produce images of higher quality, more realistic textures, and lower FID~\cite{dhariwal2021diffusion,rombach2022high,nichol2021glide} than traditional GAN~\cite{goodfellow2020generative} models. 
It gradually adds sampled Gaussian noise to an image by:
\vspace{-2pt}
\begin{equation}
q(x_t|x_{t-1}) = \mathcal{N}(x_t;\sqrt{\bar{a}_t}x_{t-1},(1-\bar{a}_t)\mathbf{I})
\label{equation:df_1}
\vspace{-2pt}
\end{equation}
where $\bar{a}_t$ is a hyperparameter and $x_t$ is the noised image at timestep $t$. 
When generating an image, the model $\theta$ carries out the denoising process over a given number of $T$ timesteps. At each timestep, the model attempts to predict the sampled Gaussian noise with the following equation:
\vspace{-2pt}
\begin{equation}
p_\theta(x_{t-1}|x_t) = \mathcal{N}(\mu_\theta(x_t), \Sigma_\theta(x_t))
\vspace{-4pt}
\label{equation:df_3}
\end{equation}

In this paper, 
our study is mainly based on three popular conditional diffusion models, i.e. DiT~\cite{peebles2022scalable}, GLIDE~\cite{nichol2021glide} and Stable Diffusion~\cite{rombach2022high}, for synthetic data generation.
These models incorporate a conditioning input $c$, which allows for the noise prediction to be conditioned on $c$ such as:
\vspace{-2pt}
\begin{equation}
p_\theta(x_{t-1}|x_t, c) = \mathcal{N}(\mu_\theta(x_t|c), \Sigma_\theta(x_t|c))
\label{equation:df_4}
\vspace{-4pt}
\end{equation}

By introducing a noise prediction network $\epsilon_\theta$ to model $\mu_\theta$, a simple mean squared error loss function is used for training the model: 
\vspace{-4pt}
\begin{equation}
\mathcal{L}_{simple}(\theta) = MSE(\epsilon_\theta(x_t), \epsilon_t)
\label{equation:df_6}
\vspace{-2pt}
\end{equation}
The objective is to minimize the distance between the predicted noise $\epsilon_\theta(x_t)$ and the GT noise $\epsilon_t$.


More specifically, to generate images with more distinctive features related to the given conditions, a classifier-free sampling strategy is employed, which encourages a high value of $p(c|x_t)$.
$p(c|x_t)$ can be represented in terms of $p(x_t|c)$ and $p(x_t)$ by using Bayes’ Rule as:
\vspace{-2pt}
\begin{equation}
p(c|x_t) = \frac{p(x_t|c) \cdot p(c)}{p(x_t)}
\vspace{-2pt}
\label{equation:df_8}
\end{equation}
By taking the logarithm and derivative on $x_t$, the following equation is obtained:
\vspace{-2pt}
\begin{equation}
\nabla_{x_t} \log p(c|x_t) \propto \nabla_{x_t}\log p(x_t|c) - \nabla_{x_t}\log p(x_t)
\label{equation:df_9}
\vspace{-2pt}
\end{equation}
Therefore, the denoising process can be directed towards removing the conditional noise by interpreting the predicted noise from the models $\epsilon_\theta$ as the score function:
\vspace{-2pt}
\begin{equation}
\hat{\epsilon}_\theta(x_t|c) = \epsilon_\theta(x_t|\emptyset) - s \cdot (\sigma_t \nabla_{x_t} \log p(c|x_t))\propto \epsilon_\theta(x_t|\emptyset) + s \cdot (\epsilon_\theta(x_t|c)-\epsilon_\theta(x_t|\emptyset))
\label{equation:df_10}
\vspace{-1pt}
\end{equation}
where $\epsilon_\theta(x_t|c)$ is the sampled noise predicted at timestep $t$ with condition $c$, and $\epsilon_\theta(x_t|\emptyset)$ is the unconditional predicted noise.
Here, the hyperparameter $s\geq1$ is used to adjust the scale of the guidance, with $s$=1 indicating that no classifier-free guidance is employed. Additionally, $\emptyset$ represents a trainable "null" condition.
Based on the performance comparison between DiT, Glide, and SD in Table~\ref{table:different_diffusion}, we default to use DiT as our synthetic data generator for knowledge distillation in this study. Data generation and student training details are presented in the supplementary material.


\begin{wraptable}{r}{0.5\textwidth}
    \begin{center}
    \vspace{-6pt}
    \resizebox{0.89\linewidth}{!}{
		  \begin{tabular}{lccc}
            \hline\noalign{\smallskip}
		  Method                         & \#Syn Images & Epoch & Acc~(\%)   \\
            \hline\noalign{\smallskip}
            GLIDE~\cite{nichol2021glide}   & 200K     &  100  & 44.58  \\ 
            SD~\cite{rombach2022high}      & 200K     &  100  & 39.95  \\ 
		  DiT~\cite{peebles2022scalable} & 200K     &  100  & 54.64  \\
            \hline
		  \end{tabular}
    }
	\end{center}
    \vspace{-4pt}
	\caption{Comparison of three state-of-the-art diffusion models on ImageNet-1K using their default hyper-parameters to generate synthetic images. "\#Syn Images" represents the total number of synthetic images. We use the pre-trained ResNet18 as the teacher to train the vanilla ResNet18 student model.}
	\label{table:different_diffusion}
\end{wraptable}

\textbf{Knowledge Distillation.}
Originally proposed by Hinton et al.~\cite{hinton2015distilling}, knowledge distillation aims to transfer the knowledge of a pretrained heavy teacher model to a lightweight student model. 
After the distillation, the student can master the expertise of the teacher and be used for final deployment. Specifically, the Kullback-Leibler~(KL) divergence loss is utilized to match the output distribution of two models, which can be simply formulated as follows:
\vspace{-1pt}
\begin{equation}
L_{kd}(q^{t}, q^{s}) = \tau^{2} KL(\sigma (q^{t}/\tau),\sigma (q^{s}/\tau)),
\label{equation:kd}
\vspace{-1pt}
\end{equation}
where $q^{t}$ and $q^{s}$ denote the logits predicted by the teacher and student. $\sigma(\cdot)$ is the softmax function and $\tau$ is the temperature hyperparameter which controls the softness of probability distribution. 



\textbf{Synthetic Data Knowledge Distillation Based on Diffusion Models~(DM-KD).}
An overview of our proposed method is illustrated in Fig.~\ref{fig:framework}. The framework consists of three models: the publicly available diffusion model, the pre-trained teacher model, and the untrained target student model. 
The diffusion model is responsible for generating specified synthetic images, denoted as $x^{n}$, based on given conditions, such as class labels or text. After synthesis, we aim to distill the teacher's knowledge to the student on the synthetic dataset.
Given the unlabeled synthetic training dataset $D=\{x^{n}\}_{n=1}^{N}$ generated by the diffusion model, we minimize the distillation loss between teacher and student models:
\vspace{-2pt}
\begin{equation}
L_{stu} =  \sum_{n}^{N}L_{kd}(f_{t}(x^{n}), f_{s}(x^{n})).
\label{equation:overview}
\vspace{-2pt}
\end{equation}
where $f_{t}$ and $f_{s}$ represent the function of teacher and student, respectively.

\section{Experiment}

\subsection{Settings.}
\textbf{Dataset.} 
The CIFAR-100~\cite{krizhevsky2009learning} dataset consists of colored natural images with $32\times32$ pixels. The train and test sets have 50K images and 10K images respectively. 
ImageNet-1K~\cite{deng2009imagenet} contains 1.28M images for training, and 50K for validation, from 1K classes. 
We also extend our method to other datasets, including ImageNet-100~(100 category), and Flowers-102~(102 category)~\cite{nilsback2006visual}. 
We use the synthetic dataset to train our student model and the real validation set to evaluate the performance.

\textbf{Data generation.} We select the 1K classes from ImageNet-1K as the default conditions for our data synthesis. 
Each category generates an equal number of images, all of which have a size of $256\times256$. For instance, in a synthesized dataset of 200K images, every category contains 200 images.
To conduct our knowledge distillation experiments on ImageNet-1K, ImageNet-100, and Flowers-102 datasets, we resize the synthetic dataset to $224\times224$. For CIFAR-100, the synthetic training dataset contains 50K images that are resized to $32\times32$.

\textbf{Implementation details.} 
All of the experiments are implemented by Pytorch~\cite{paszke2019pytorch} and conducted on a server containing eight NVIDIA RTX 2080Ti GPUs.
We follow the same training schedule as previous knowledge distillation methods~\cite{tian2019contrastive,chen2021distilling}.
We use the stochastic gradient descents~(SGD) as the optimizer with momentum 0.9 and weight decay 5e-4. 
For CIFAR-100, the initial learning rate is 0.05 and divided by 10 at 150, 180, and 210 epochs, for a total of 240 epochs. The mini-batch size is 64. 
For ImageNet-1K, the weight decay is 1e-4 and the batch size is 256. The initial learning rate is set to 0.1 and divided by 10 at 30, 60, and 90 epochs, for a total of 100 epochs. We set the temperature $\tau$ to 10 by default. For each dataset, we report the Top-1 classification accuracy. The results are averaged over 3 runs. More training details are presented in the supplementary material.

\subsection{Comparison to existing synthesis-based distillation methods.}

\begin{table}[t]
	\begin{center}
		\resizebox{1\linewidth}{!}
		{
                \begin{tabular}{ccccccc}
                \hline\noalign{\smallskip}
                \multirow{2}*{Method}  & \multirow{2}*{Syn Method} & T:ResNet34 & T:VGG11 & T:WRN40-2 & T:WRN40-2 & T:WRN40-2 \\
                ~                      & ~            & S:ResNet18 & S:ResNet18 & S:WRN16-1 & S:WRN40-1 & S:WRN16-2 \\
                \hline\noalign{\smallskip}
                Teacher                & -            & 78.05 & 71.32 & 75.83  & 75.83  & 75.83   \\
                Student                & -            & 77.10 & 77.10 & 65.31  & 72.19  & 73.56   \\
                KD                     & -            & 77.87 & 75.07 & 64.06  & 68.58  & 70.79   \\
                \hline\noalign{\smallskip}
                DeepInv~\cite{yin2020dreaming} & Inversion    & 61.32 & 54.13 & 53.77  & 61.33  & 61.34 \\
                DAFL~\cite{chen2019data}       & Inversion    & 74.47 & 54.16 & 20.88  & 42.83  & 43.70 \\
                DFQ~\cite{choi2020data}        & Inversion    & 77.01 & 66.21 & 51.27  & 54.43  & 64.79 \\
                FastDFKD~\cite{fang2022up}     & Inversion    & 74.34 & 67.44 & 54.02  & 63.91  & 65.12 \\
                \hline\noalign{\smallskip}
                One-Image~\cite{asano2023the}  & Augmentation & 74.56 & 68.51 & 34.62  & 52.39  & 54.71 \\
                DM-KD~(Ours)                           & Diffusion    & \textbf{76.58} & \textbf{70.83} & \textbf{56.29}  & \textbf{65.01}  & \textbf{66.89} \\
                \hline
			\end{tabular}
		}
	\end{center}
	\caption{Student accuracy on CIFAR-100 validation set. }
	\label{table:cifar}
        \vspace{-5pt}
\end{table}

\begin{table}[t]
	\begin{center}
		\resizebox{0.9\linewidth}{!}
		{
                \begin{tabular}{cccccc}
                \hline\noalign{\smallskip}
                \multirow{2}*{Method}          & \multirow{2}*{Syn Method} & \multirow{2}*{Data Amount} & \multirow{2}*{Epoch}& T: ResNet50/18 & T: ResNet50/18   \\
                ~                              & ~                        & ~     &  ~   & S: ResNet50 & S: ResNet18 \\
                \hline\noalign{\smallskip}
                Places365+KD                   & -                        & 1.8M  & 200  & 55.74 & 45.53       \\
                BigGAN~\cite{brock2018large}   & GAN                      & 215K  & 90   & 64.00 & -           \\
                DeepInv~\cite{yin2020dreaming} & Inversion                & 140K  & 90   & 68.00 & -           \\
                FastDFKD~\cite{fang2022up}     & Inversion                & 140K  & 200  & 68.61 & 53.45       \\
                \hline\noalign{\smallskip}
                One-Image~\cite{asano2023the}  & Augmentation             & 1.28M & 200  & 66.20 & -           \\
                \hline\noalign{\smallskip}
                \multirow{3}*{DM-KD~(Ours)}    & \multirow{3}*{Diffusion} & 140K  & 200 & 66.74 & 60.10  \\
                ~                              & ~                        & 200K  & 200 & 68.63 & 61.61  \\ 
                ~                              & ~                        & 1.28M & 200 & \textbf{72.43} & \textbf{68.25}  \\ 
                \hline   
			\end{tabular}
		}
	\end{center}
	\caption{Student accuracy on ImageNet-1K validation set. 
 }
	\label{table:imagenet}
        \vspace{-15pt}
\end{table}

\textbf{Data-free Distillation.}
Existing data-free distillation methods are all based on the white-box teacher model for distillation. These methods primarily utilize information within the white-box teacher model, such as layer statistics, to generate samples and construct training sets that approximate the original data for distillation.
Previous methods have three limitations. Firstly, it requires careful of design of the generative method, which is complex and time-consuming. Secondly, it becomes ineffective when the white-box teacher model is not available and only predictions through APIs are provided. Thirdly, current methods face difficulties in scaling with larger data volumes due to the limited diversity of synthetic data~\cite{luo2020large,choi2020data,fang2021contrastive}. 
Our method effectively solves the above problems. By adopting the publicly available advanced diffusion model, samples can also be generated when the teacher is a black-box model. At the same time, the large-scale diffusion model can easily generate a large number of diverse high-resolution samples for distillation.
In Tables~\ref{table:cifar} and~\ref{table:imagenet}, we compare our method with mainstream data-free methods, and our method shows very competitive performance when trained on the same amount of synthetic data. By simply introducing more synthetic training samples, our method significantly improves the performance of data-free distillation by a large margin, as shown in Table~\ref{table:imagenet}.

\textbf{One Image Distillation.}
One-Image-Distill~\cite{asano2023the} first performs multiple random crops on a large image~(i.e., $2560\times1920$), and then applies data augmentation techniques to the cropped images to create a synthetic image set. The synthetic dataset contains approximately 50K images for CIFAR-100 and 1.28M images for ImageNet-1K.
It's critical to carefully select the source image for One-Image-Distill since sparse images will perform much worse than dense images as mentioned in the original paper. However, our method doesn't require such detailed selection operations. We can directly generate images through the diffusion model given the target label space.
As shown in Table~\ref{table:cifar} and Table~\ref{table:imagenet}, our method shows better performance for the same or larger data volume.
Note that the results in Table~\ref{table:cifar} were not reported in the original paper, so we adopt the "Animals" image and reimplement the method based on the official code\footnote{https://github.com/yukimasano/single-img-extrapolating}.


\textbf{Extension to other datasets.} 
Our synthetic dataset is generated based on the 1K classes of ImageNet-1K. To verify the generalizability of our method, we extended it to other datasets, including CIFAR-100~\cite{krizhevsky2009learning}, ImageNet-100~\cite{deng2009imagenet}, and Flowers-102~\cite{nilsback2006visual}. Specifically, the teacher pre-trains on the specified real dataset and then performs knowledge distillation based on our synthetic dataset. The results are reported in Tables~\ref{table:cifar} and~\ref{table:other_dataset}, and the excellent performance on these three datasets indicates our method demonstrates good generalization to other datasets. Notably, it is surprising to find that our method achieved a great distillation performance on the fine-grained Flowers-102 dataset, even though the synthetic dataset categories do not intersect with the Flowers-102 dataset categories.
There are two possible reasons for such a good generalization. First, the 1K classes of the synthetic data contain most of the common classes, enabling students to learn robust general features during training. Second, in knowledge distillation, the influence of data domain shift on students can be effectively weakened by the supervision of the pretrained teacher model.

\begin{figure*}[t]
        \hspace{20pt}
	\includegraphics[width=0.8\linewidth]{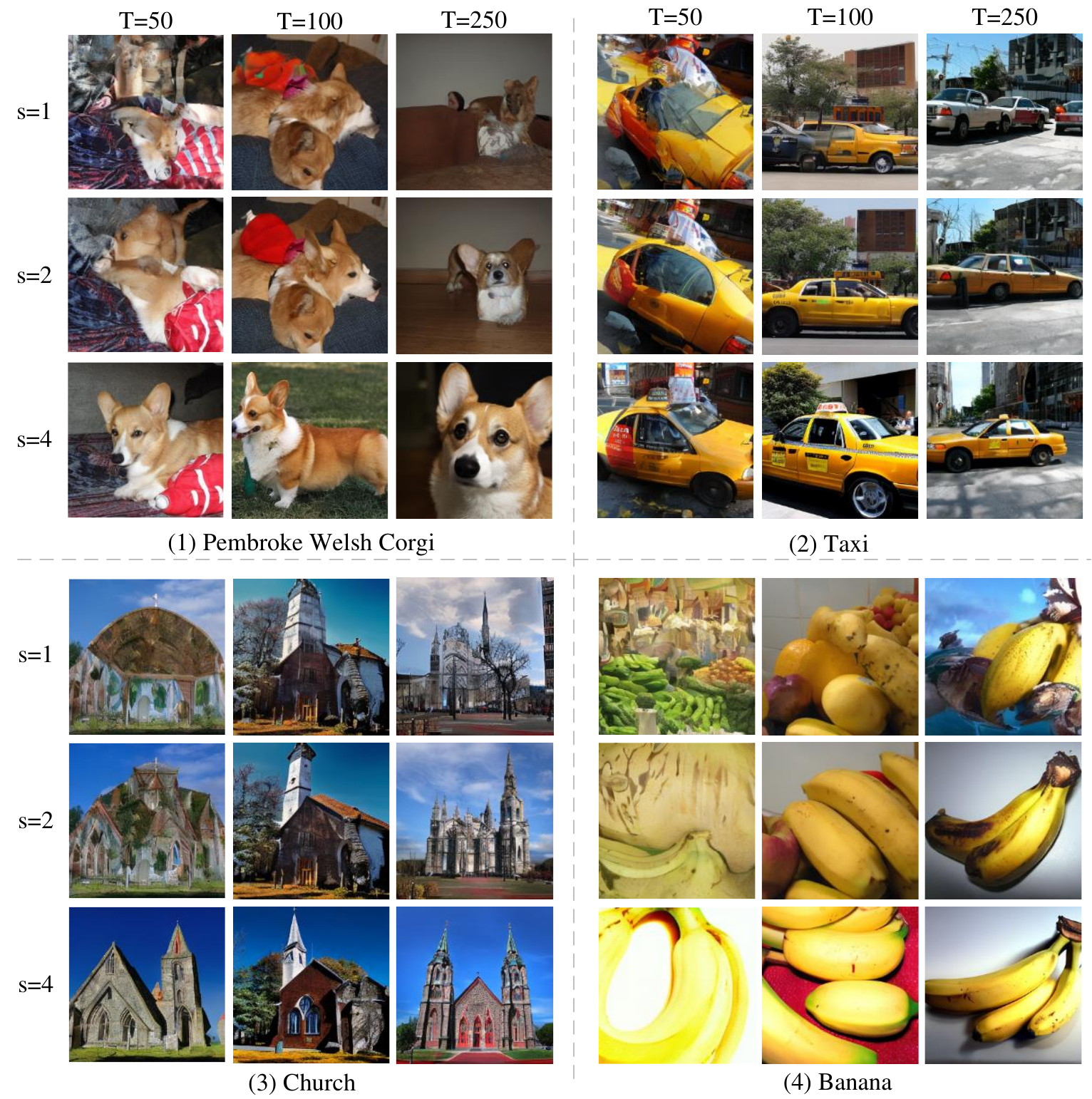}
        \vspace{-4pt}
	\caption{Visualized fidelity of synthetic images with different scaling factors $s$ and sampling steps $T$. Increasing the value of $s$ or $T$ results in higher fidelity images. The example images generated with $s$=2 contain unrealistic elements, such as a dog with two heads or a church building that is unnaturally distorted, and their fidelity and coherence are much lower than those generated with $s$=4.}
	\label{fig:Visualization}
        \vspace{-10pt}
\end{figure*}

\begin{table}[h]
	\begin{center}
		\resizebox{0.77\linewidth}{!}
		{
                \begin{tabular}{ccccc}
                \hline\noalign{\smallskip}
                \multirow{2}*{Datasets}  & Categories    & Teacher   & One-Image~\cite{asano2023the}  & Ours    \\
                ~                        & (\#Classes)   & ResNet18  & ResNet50   & ResNet50  \\
                \hline\noalign{\smallskip}
                ImageNet-100~\cite{deng2009imagenet}  & Objects~(100)      & 89.6  & 84.4 & \textbf{85.9} \\
                \hline\noalign{\smallskip}
                Flowers-102~\cite{nilsback2006visual} & Flower types~(102) & 87.9  & 81.5 & \textbf{85.4} \\
                \hline
			\end{tabular}
		}
	\end{center}
	\caption{Student accuracy on ImageNet-100, and Flowers-102 datasets. Our DM-KD demonstrates good generalization to other datasets. Notably, even when there is no intersection of categories between the synthetic dataset and the Flowers-102 dataset, our method still achieves high performance.}
	\label{table:other_dataset}
\vspace{-10pt}
\end{table}

\subsection{Low-fidelity synthetic images are better teaching materials.}

When synthesizing images, two hyperparameters determine the overall fidelity of the synthetic images: the value of the classifier-free guidance scaling factor $s$, and the total number of sampling steps $T$. 
In general, the parameter $s$ controls the strength of the injection of conditional distinction features.n A higher value of $s$ produces an image with richer conditional features and higher image fidelity. Conversely, a lower value of $s$~(where $s$=1 means no guidance) produces an image with lower fidelity, sometimes, even with image distortion and deformation.
Another parameter, $T$, determines the number of denoising timesteps carried out during image generation. Typically, a larger value of $T$ results in a more detailed image with higher fidelity, but it also increases the generation time. DiT uses $s$=4 and $T$=250 as its default parameters for high-fidelity image synthesis.

To investigate the relationship between synthetic fidelity and distillation performance, datasets are synthesized using the DiT model with varying classifier-free guidance scaling factors~($s$) and sampling steps~($T$). 
Examples of different categories from the synthesized datasets with different scaling factors $s$ and sampling steps $T$ are visualized in Fig.~\ref{fig:Visualization}.
In this paper, the fidelity of the synthesized datasets is evaluated using the ImageReward~\cite{xu2023imagereward} metric, which is the state-of-the-art in synthesis dataset evaluation. The ImageReward metric scores are presented in Table ~\ref{table:imagereward}, where higher scores indicate higher fidelity and human preference. In this section, both teacher and student models adopt the ResNet18 model. For the teacher model, we use the pre-trained weights on torchvision~\footnote{https://pytorch.org/vision/main/models.html}. The experiments are conducted on 200K synthetic datasets and the students are trained by 100 epochs for experimental efficiency.


\begin{minipage}{\textwidth}
\begin{minipage}[t]{0.50\textwidth}
    \begin{center}
        \makeatletter\def\@captype{table}\makeatother
        \resizebox{0.94\linewidth}{!}
        {
                \begin{tabular}{cccccc}
                \hline\noalign{\smallskip}
                \multirow{2}*{\makecell{Scaling\\Factor $s$}} & \multicolumn{5}{c}{Sampling Step $T$}     \\
                \noalign{\smallskip}
                    & 50    & 100   & 150   & 200   & 250   \\
                \hline\noalign{\smallskip}
    		  1   & -1.087 & -1.010 & -0.984 & -0.966 & -0.948 \\
                2   & \cellcolor{lightgray!30}-0.332 & \cellcolor{lightgray!30}-0.292 & \cellcolor{lightgray!30}-0.279 & \cellcolor{lightgray!30}-0.270 & \cellcolor{lightgray!30}-0.264 \\ 
                3   & -0.112 & -0.093 & -0.086 & -0.086 & -0.076 \\
                4   & -0.026 & -0.016 & -0.011 & -0.007 & -0.003 \\
                \hline
			\end{tabular}
        }
         \captionsetup{font={small}}
        \caption{ ImageReward~\cite{xu2023imagereward} with different classifier-free guidance scaling factors $s$ and sampling steps $T$. A larger value denotes higher image fidelity.}
        \label{table:imagereward}
    \end{center}
\end{minipage}
\hspace{5pt}
\begin{minipage}[t]{0.47\textwidth}
    \begin{center}
    \makeatletter\def\@captype{table}\makeatother
        \resizebox{0.94\linewidth}{!}
        {
                \begin{tabular}{cccccc}
                \hline\noalign{\smallskip}
                \multirow{2}*{\makecell{Scaling\\Factor $s$}} & \multicolumn{5}{c}{Sampling Step $T$}     \\
                \noalign{\smallskip}
                    & 50    & 100   & 150   & 200   & 250   \\
                \hline\noalign{\smallskip}
    		  1   & 52.46 & 54.44 & 54.50 & 55.03 & 54.90  \\
                2   & \cellcolor{lightgray!30}56.88 & \cellcolor{lightgray!30}\textbf{57.22} & \cellcolor{lightgray!30}57.12 & \cellcolor{lightgray!30}57.14 & \cellcolor{lightgray!30}57.04  \\
                3   & 56.02 & 56.12 & 56.28 & 56.25 &  56.09      \\
                4   & 54.92 & 54.78 & 54.95 & 54.81 &  \underline{54.64}  \\
                \hline
			\end{tabular}
        }
         \captionsetup{font={small}}
        \caption{ Student accuracy with different scaling factors $s$ and sampling steps $T$. Student models are trained for 100 epochs.}
        \label{table:factor_step_ratio}
    \end{center}
    \end{minipage}
\end{minipage}

\begin{table}[h]
	\begin{center}
		\resizebox{0.9\linewidth}{!}
		{
                \begin{tabular}{cccccc}
                \hline\noalign{\smallskip}
                Teacher        & ResNet18 & ResNet34 & VGG16 & ResNet50 & ShuffleV2\\
                Acc.~(\%)            & 69.75    & 73.31    & 73.36 & 76.13    & 69.36 \\
                \hline\noalign{\smallskip}
                Student                        & ResNet18 & ResNet18 & VGG11  & ResNet34 & ResNet50 \\
                Low Fidelity~($s$=2, $T$=100, reward=-0.292)  & 57.22    & 54.17    & 51.81  & 57.78    & 63.66    \\
                High Fidelity~($s$=4, $T$=250, reward=-0.003) & 56.19    & 49.73    & 47.51  & 52.24    & 56.20     \\
                \hline
			\end{tabular}
		}
	\end{center}
	\caption{Distillation performance for different teacher-student pairs under low- and high-fidelity synthetic images. The low-fidelity images are more effective for various teacher-student pairs. }
	\label{table:low_high_fidelity_exp}
        \vspace{-8pt}
\end{table}

\textbf{Results.} 
For image classification tasks, it is commonly believed that classification models will demonstrate better generalization ability on real data when high-fidelity, photo-realistic synthesized images are utilized as the training dataset. 
Existing data-free distillation methods~\cite{chen2019data,yin2020dreaming} follow a similar idea by synthesizing realistic datasets for distillation. 
However, we find that the distillation performance of images synthesized with default parameters is suboptimal. 
To assess the distillation performance achieved with different synthetic datasets, we report the student accuracy in Table ~\ref{table:factor_step_ratio} and Table~\ref{table:low_high_fidelity_exp}, which correspond to the datasets in Table~\ref{table:imagereward}. 
Our study shows that high-fidelity images, as indicated by a high ImageReward~\cite{xu2023imagereward} score, generated with default parameters, exhibit weaker distillation performance than low-fidelity ones. By progressively decreasing the values of both the scaling factor $s$ and sampling step $T$, a better distillation performance can be achieved. These findings suggest that low-fidelity images are more effective as learning materials for students during the distillation process.
In addition, when setting $s$=1, which means that there is no classifier-free guidance in the image generation process, a significant drop in student performance is observed. This suggests that poor fidelity generated images may hinder the student model's ability to learn effectively in logit representation. 
Our experiments show that setting $s$=2 and using a sampling step of 100 can generate images with relatively low fidelity, which results in the best performance for ImageNet-1K knowledge distillation (see in Table~\ref{table:factor_step_ratio} and Table~\ref{table:low_high_fidelity_exp}).

\textbf{Scaling with more synthetic data.} 
Based on the above findings, we proceed to evaluate the distillation performance on large-scale data by generating synthetic images of varying data amounts, ranging from 100K to 2.0M. In the following experiments, we use a scaling factor of $s$=2 and sampling step $T$=100 to construct the synthetic dataset, unless otherwise specified. Fig.~\ref{fig:acc_data_amount} shows that the performance continues to improve as the number of synthetic images increases up to 2.0M. As the data amount increases, we observe that the performance improvement begins to plateau.

\begin{figure*}[t]
	\centering
        \includegraphics[width=0.32\linewidth]{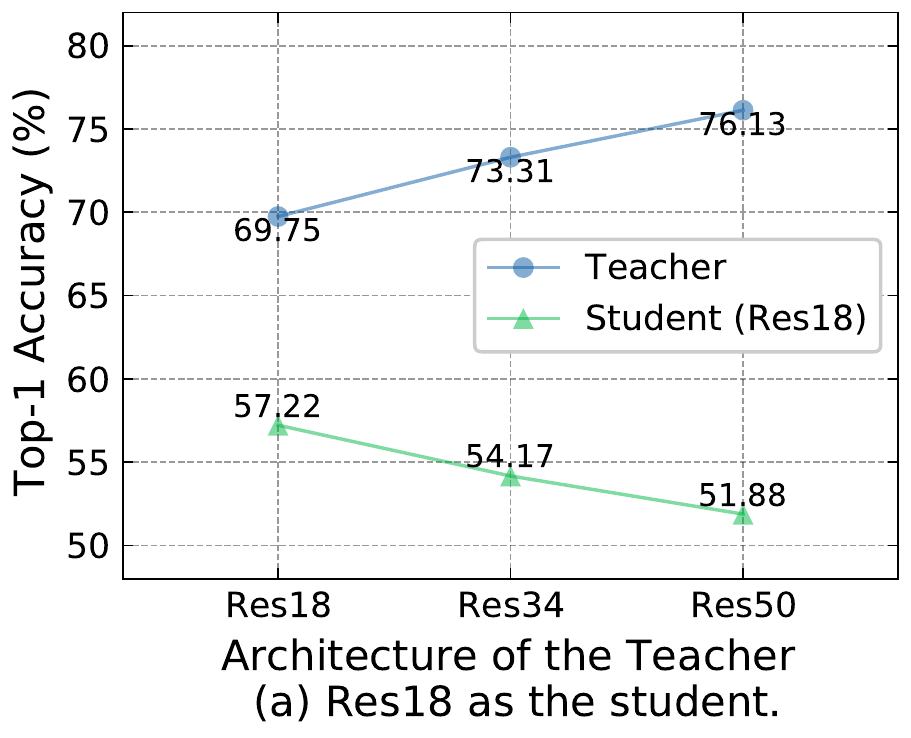}
        \includegraphics[width=0.32\linewidth]{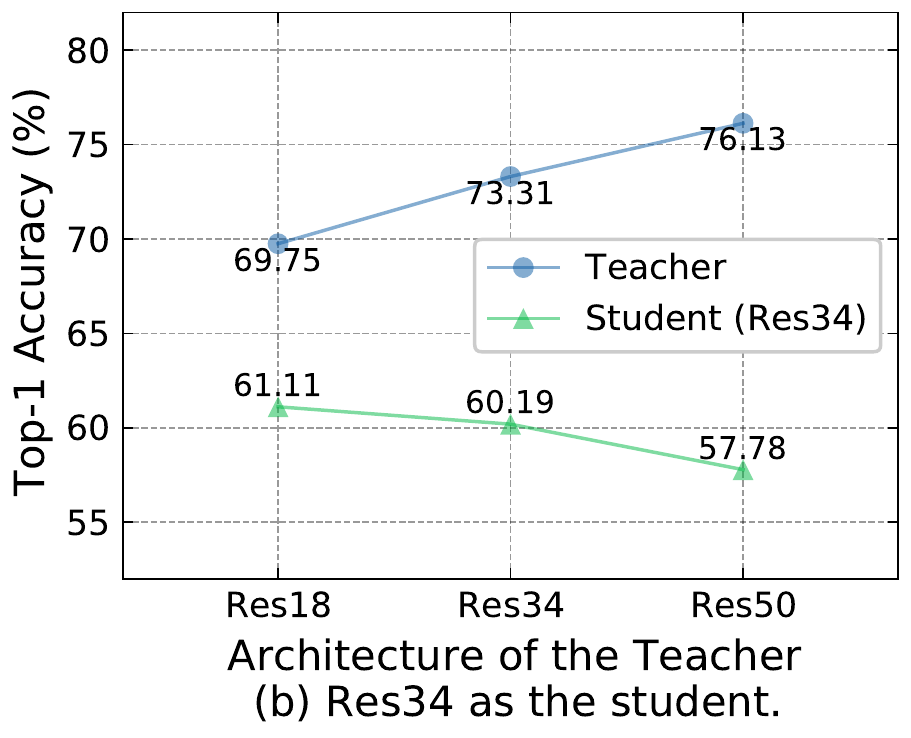}
        \includegraphics[width=0.32\linewidth]{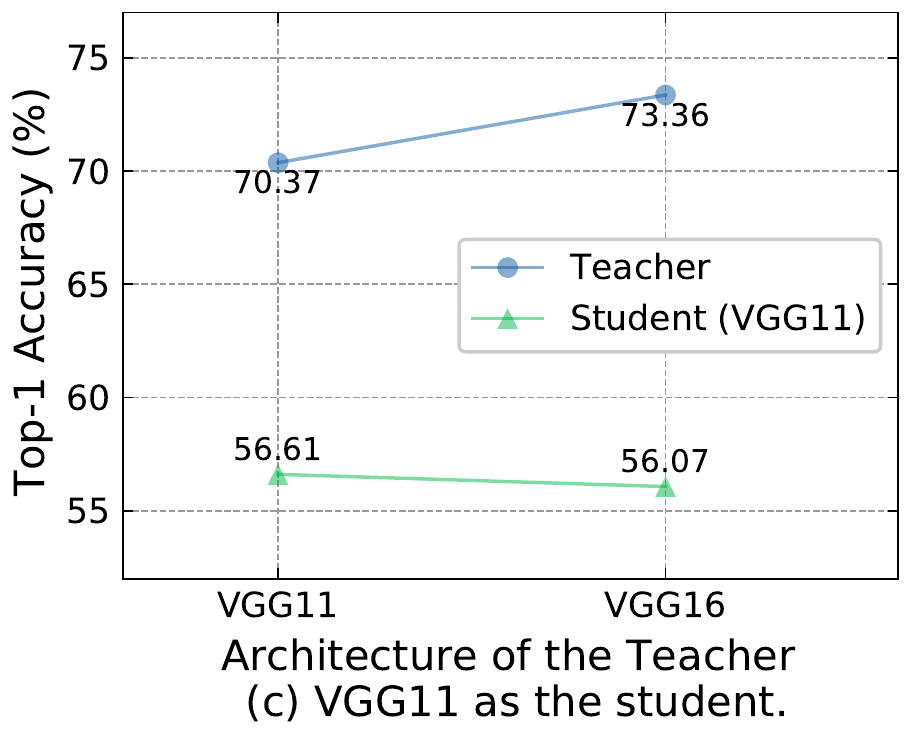}
        \vspace{-2pt}
        
	\caption{Top-1 classification accuracy~(\%) of different teacher-student pairs on ImageNet dataset. Relatively weak classifiers bring better distillation performance. Student models are trained for 100 epochs.}
	\label{fig:weak_cls}
        \vspace{-4pt}
\end{figure*}

\begin{table}[h]
	\begin{center}
		\resizebox{0.8\linewidth}{!}
		{
			\begin{tabular}{cccc|ccc|cc}
                \hline\noalign{\smallskip}
				Teacher  & SNV2-0.5 & Res18 & Res34 & SNV2-0.5 & Res18 & Res34 & SNV2-0.5 & Res18 \\
                Acc.~(\%) & 60.55    & 69.75 & 73.31 & 60.55    & 69.75 & 73.31 & 60.55    & 69.75 \\
                \hline\noalign{\smallskip}
				Student  & Res18    & Res18 & Res18 & Res34    & Res34 & Res34 & Res50    & Res50 \\
				SynKD    & 54.03    & 57.22 & 54.17 & 56.83    & 61.11 & 60.19 & 58.11    & 62.74 \\
                \hline
			\end{tabular}
		}
	\end{center}
	\caption{Knowledge distillation with large teacher-student capacity gaps. A relatively weak teacher with a small teacher-student gap generally leads to better performance.}
	\label{table:large_gap}
        \vspace{-10pt}
\end{table}




\begin{wraptable}{r}{0.55\textwidth}
		\resizebox{0.92\linewidth}{!}{
		\begin{tabular}{ccccc}
            \hline\noalign{\smallskip}
		  \#Syn Images                   & 50K   & 100K  & 200K  & 400K  \\
		  Train Epoch                    & 400   & 200   & 100   & 50    \\
            \hline\noalign{\smallskip}
		  ResNet18$\rightarrow$ResNet18  & 54.84 & 55.00 & 57.22 & \textbf{57.69} \\
            \hline
		\end{tabular}
        }
	\caption{Data diversity. Generating more data samples increases the diversity of the synthetic dataset, leading to better distillation performance.}
	\label{table:data_diversity}
        \vspace{-10pt}
\end{wraptable}

\textbf{Data diversity.}
In this experiment, we aim to validate whether generating more samples brings greater diversity to the dataset and thus leads to better distillation performance. To achieve this, we fix the number of total training iterations (i.e., Data Amount$\times$Train Epochs) and scale the training schedule based on the data volume. Our results, presented in Table~\ref{table:data_diversity}, demonstrate that generating more data increases the diversity in the synthetic dataset, resulting in improved performance.


\subsection{Relatively weak classifiers are better teachers.}

\textbf{Experiment setup.} To validate if relatively weak classifiers are better teachers, we carefully select multiple teacher-student model pairs with different architectures, including ResNet~\cite{he2016deep}, VGG~\cite{simonyan2014very}, and ShuffleNetV2~(SNV2)~\cite{ma2018shufflenet}. The experiments are conducted on 200K synthetic datasets and the students are trained by 100 epochs.

\begin{figure}[t]
  \begin{minipage}[t]{0.46\linewidth}
    \centering
    \includegraphics[width=0.84\linewidth]{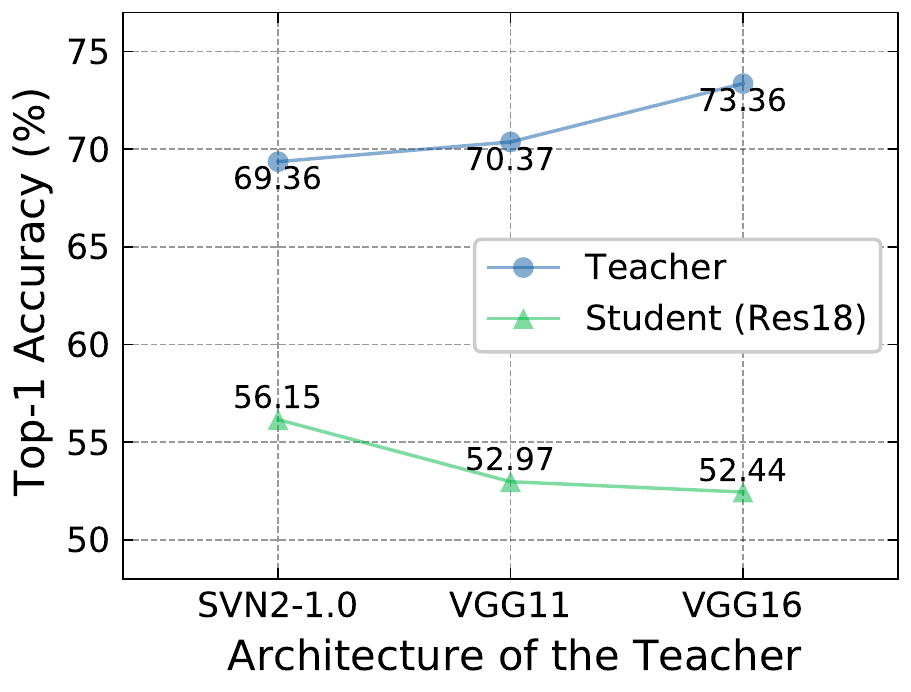}
    \captionsetup{font={small}}
    \vspace{-4pt}
    \caption{Top-1 classification accuracy~(\%) on ImageNet dataset. Teachers and students are different architectures. Students are trained for 100 epochs.}
    \label{fig:weak_cls_2}
  \end{minipage}%
  \hspace{10pt}
  \begin{minipage}[t]{0.52\linewidth}
    \centering
    \includegraphics[width=0.9\linewidth]{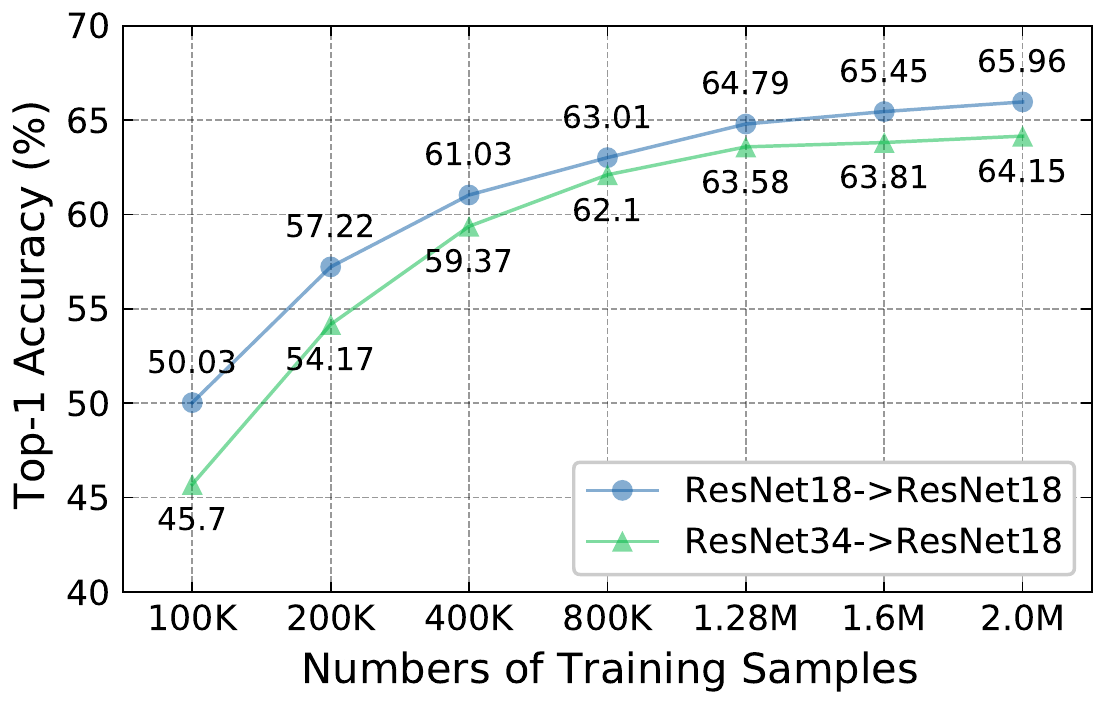}
     \captionsetup{font={small}}
    \vspace{-4pt}
    \caption{Improved classification accuracy of the student model with increasing numbers of synthetic images used for distillation. Student models are trained for 100 epochs.}
    \label{fig:acc_data_amount}
  \end{minipage}
\vspace{-10pt}
\end{figure}

\textbf{Results.}
When training on real datasets, it is common to use a relatively large teacher model to train the student model, such as distilling ResNet34 to ResNet18. In general, smaller teacher models often fail to achieve satisfactory distillation performance compared to larger teacher models. However, when working with synthetic datasets, we observe the \textbf{opposite} phenomenon: relatively weak teacher models can actually achieve better distillation performance than strong ones, as shown in Fig.~\ref{fig:weak_cls} and Fig.~\ref{fig:weak_cls_2}. 
Interestingly, we found that as the capacity of the teacher model increases, a significant drop in performance is observed.
Specifically, when training ResNet34 on the synthetic dataset, using ResNet18 as the teacher model leads to a 3\% improvement in performance compared to using ResNet50 as the teacher model.
These results highlight the importance of carefully selecting a teacher model when performing knowledge distillation and suggest that using a smaller, weaker teacher model may be preferable when working with synthetic datasets.

\textbf{Should teachers be as weak as possible?}
To answer this question, we conduct a series of experiments using three groups of teacher-student pairs with large capacity gaps, as shown in Table.~\ref{table:large_gap}. 
We choose SNV2-0.5 as the teacher to test the effect that the teacher is obviously weaker than the student.
Our results show that when the teacher-student gap is large, it does not necessarily lead to better performance. In fact, our experiments suggest that using a relatively weak teacher model may be a better choice for optimizing knowledge transfer.

\subsection{Ablation Study}

\begin{wrapfigure}{r}{0.4\linewidth}
        \centering
	\includegraphics[width=\linewidth]{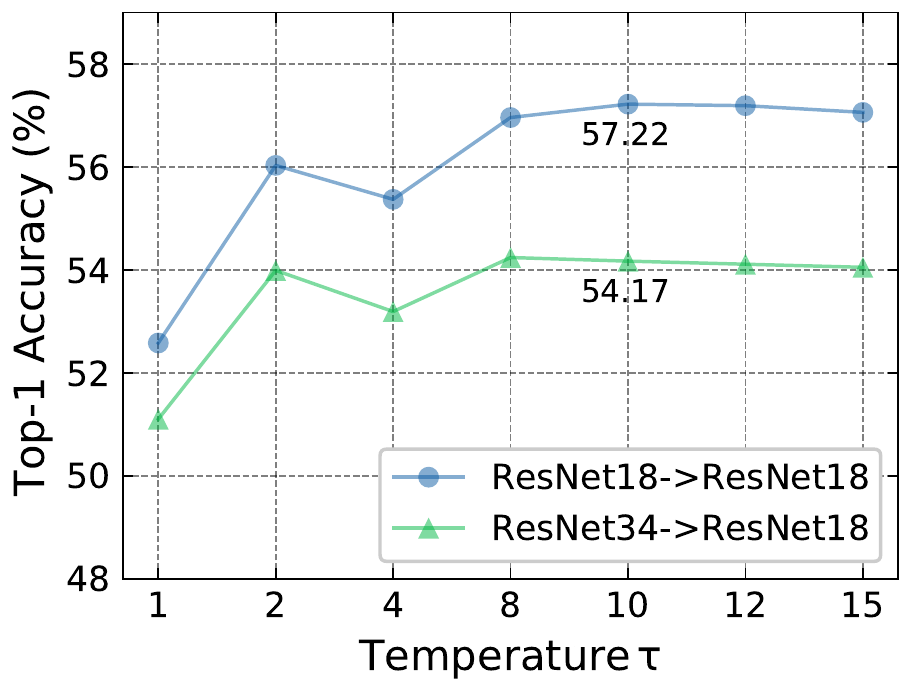}
	\caption{Grid search for values of the temperature hyperparameter. The best performance is achieved when $\tau$=10.}
        \vspace{-10pt}
	\label{fig:temp_values}
\end{wrapfigure}

\textbf{Temperature hyperparameter.}
As discussed in previous works~\cite{hinton2015distilling,chandrasegaran2022revisiting,li2022curriculum}, the temperature parameter is a crucial factor in controlling the smoothness of probability distributions. It determines the level of difficulty involved in the process and is an essential component in achieving optimal results. 
The temperature parameter $\tau$ has been set to different values in previous works (e.g., 3 in DeepInv~\cite{yin2020dreaming}, 8 in One-Image~\cite{asano2023the}, 20 in FastDFKD~\cite{fang2022up}). In Fig.~\ref{fig:temp_values}, we present the detailed distillation results obtained under different temperature values. The best result is obtained when we set $\tau$=10.


\textbf{Training with hard labels.}
The DiT model uses class labels as input to generate images, making it possible to use these labels as hard labels to supervise the model's training. In order to investigate the potential benefits of hard label supervision for synthetic datasets, we conduct experiments as presented in Table~\ref{table:hard_soft_label}. 
The experiments involved training the student model with soft labels only, hard labels only, and a combination of hard and soft labels. 
For the joint hard label and soft label training, we follow the traditional distillation methods~\cite{hinton2015distilling,zhang2018deep,zhao2022decoupled} to weights the two losses at a 1:1 ratio.
The results indicate that utilizing hard labels during training actually leads to worse performance compared to using only soft labels. This finding confirms the existence of a domain shift between the synthetic and real datasets, as mentioned in~\cite{he2022synthetic}. However, by using the distillation method, the impact of the domain shift is largely reduced.

\begin{table}[h]
	\begin{center}
	\resizebox{0.65\linewidth}{!}
        {
		      \begin{tabular}{cccccc}
			\hline\noalign{\smallskip}
			\multirow{2}*{Hard Label} & \multirow{2}*{Soft Label} & T: ResNet18 & T: ResNet34 & T: VGG16 \\
			    ~                         & ~                         & S: ResNet18 & S: ResNet18 & S: VGG11 \\
			\hline\noalign{\smallskip}
			                             & \checkmark                & \textbf{57.22} & \textbf{54.17} & \textbf{51.81}   \\
			\checkmark                &                           & 42.40       & 42.58       & 41.10   \\
			\checkmark                & \checkmark                & 56.08       & 53.61       & 50.32     \\
			\hline
		      \end{tabular}
		}
	\end{center}
	\caption{Accuracy comparison of hard and soft labels. Soft labels work better than hard labels and both.}
	\label{table:hard_soft_label}
        \vspace{-10pt}
\end{table}


\section{Conclusion}
In this paper, we systematically study whether and how synthetic images generated from state-of-the-art diffusion models can be used for knowledge distillation without access to real data. 
Through our research, we have three key findings: (1)~extensive experiments demonstrate that synthetic data from diffusion models can easily achieve state-of-the-art performance among existing synthesis-based distillation methods, (2)~low-fidelity synthetic images are better teaching materials for knowledge distillation, and (3)~relatively weak classifiers are better teachers.


\textbf{Limitations and future work.}
Due to limited computing resources, we were not able to conduct experiments on large models (e.g., ViT~\cite{dosovitskiy2020image}, Swin Transformer~\cite{liu2021swin}) with larger data volumes.
Besides, this paper focuses on the original sampling manner of the considered diffusion models and does not discuss the influence of the advanced sampling manners, e.g., DPM-Solver~\cite{lu2022dpm}, DPM++~\cite{lu2022dpmplus}. We plan to discuss them in our future work.

\section{Acknowledgement}
This work was supported by the Young Scientists Fund of the National Natural Science Foundation of China~(Grant No.62206134) and the Tianjin Key Laboratory of Visual Computing and Intelligent Perception (VCIP). Computation is supported by the Supercomputing Center of Nankai University (NKSC).

{\small
 \bibliographystyle{plain}
 \bibliography{nips23}
}

\clearpage
\appendix
\section{Appendix}

\subsection{Data generation.}

To generate the desired images, we utilize off-the-shelf text-to-image diffusion models such as GLIDE~\cite{nichol2021glide} and Stable Diffusion~\cite{rombach2022high}. The text prompt used in these models has a fixed format of "A realistic photo of {\textit{class}}," where "\textit{class}" specifies the category of the target image.
When using GLIDE, we employ the official default settings for inference, which consist of a sampling step of 100 and a classifier-free guidance scale of 3.0. Similarly, for Stable Diffusion, we utilize the official settings of a sampling step of 50 and a classifier-free guidance scale of 7.5. 


DiT~\cite{peebles2022scalable} differs from the text-to-image method in that it uses class labels as input to generate synthetic images. To generate an image corresponding to a specific class in the ImageNet-1K dataset, we input the index of that class. For example, to generate an image of a Pembroke Welsh Corgi, we would input index 263.

\subsection{Training details.}
\textbf{CIFAR-100.} We first resize the synthesized image of size $256\times256$ to $32\times32$. In addition to applying the classic distillation data augmentation scheme, i.e., random cropping and horizontal flipping, we use the CutMix~\cite{yun2019cutmix} augmentation method during training, in order to align with the One-Image~\cite{asano2023the} augmentation scheme. The temperature $\tau$ is set to 10.

\textbf{ImageNet-1K.} We have two training schedules in this paper. The first is the classic distillation training strategy, which is to train 100 epochs and divide the learning rate by 10 in the 30th, 60th, and 90th epochs. We use this as the default training strategy unless otherwise stated. In Table~\ref{table:imagenet}, we adopted the second training strategy, which is the same as that of FastDFKD~\cite{fang2022up}. This involves training for 200 epochs, with the learning rate divided by 10 at the 120th, 150th, and 180th epochs. 
Specifically, we choose the teacher model with the same structure as the student for distillation in Table~\ref{table:imagenet}. In the 5th column, we use ResNet50 as the teacher to train the student ResNet50, while in the 6th column, we use ResNet18 as the teacher to train the student ResNet18. 
For data augmentation, following the  setting of One-Image~\cite{asano2023the}, we adopt the CutMix method during training.

\subsection{Analysis.}

When using DiT to generate images, due to the classifier-free guidance mechanism, the generated images will have a distinct appearance that aligns with the specific class. This allows the classifier to classify these generated images more easily than real images, resulting in higher class confidence, as shown in Table~\ref{table:teacher_on_syn}.
However, this high confidence in the synthesized images can also result in a sharp output distribution, making it challenging for knowledge distillation to effectively transfer knowledge of class similarity. A smooth target distribution would be more effective for knowledge transfer.

\begin{table}[h]
	\begin{center}
        \resizebox{0.5\linewidth}{!}
        {
            \begin{tabular}{cccccc}
            \hline\noalign{\smallskip}
            \multirow{2}*{\makecell{Sampling\\Factor $s$}} & \multicolumn{5}{c}{Sampling Step $T$}     \\
            \noalign{\smallskip}
                & 50    & 100   & 150   & 200   & 250  \\
            \hline\noalign{\smallskip}
    	1   & 44.68 & 48.62 & 50.53 & 50.40 & 50.83 \\
            2   & \cellcolor{lightgray!30}80.31 & \cellcolor{lightgray!30}81.45 & \cellcolor{lightgray!30}81.72 & \cellcolor{lightgray!30}82.07 & \cellcolor{lightgray!30}81.90 \\
            3   & 87.56 & 87.73 & 87.46 & 87.53 & 87.53 \\
            4   & 88.96 & 89.00 & 88.72 & 88.76 & 88.71 \\
            \hline
		\end{tabular}
        }
        \end{center}
        \caption{Pretrained ResNet18 teacher evaluated on synthetic dataset with different scaling factors $s$ and sampling steps $T$. When classifier-free guidance is used~(i.e., $s>1$), these results are significantly higher than the accuracy of 69.75\% on the real ImageNet validation set. When $s$=1, we can observe that the accuracy of the pre-trained teacher is only around 50\%. This implies that the teacher is unable to provide accurate soft labels for distillation.}
        \label{table:teacher_on_syn}
        \vspace{-10pt}
\end{table}

Both our second and third findings in Section~\ref{section:intro} are attempts to reduce the sharpness of the distribution and create a smooth learning target for distillation. 
(1) The goal of creating low-fidelity samples is to increase the classification difficulty of the classifier and decrease its tendency to become overconfident in predicting a certain class. 
As shown in Table~\ref{table:variance}, by gradually reducing the scaling factor $s$, the variance is gradually reduced, which means that the distribution generated by the teacher becomes smoother. The best performance is achieved when we set $s$=2, which corresponds to the lowest scaling factor in the table.
(2) Weak classifiers are less discriminative compared to strong classifiers and tend to produce smoother outputs. In Table~\ref{table:variance}, we compare the output variance of pretrained ResNet18 and ResNet34 teachers on the synthetic dataset. Our results indicate that for different scaling factors $s$, ResNet18 consistently achieves a lower variance than ResNet34. This shows that ResNet18 can produce smoother output for distillation.

\begin{table}[h]
    \begin{center}
	\resizebox{0.4\linewidth}{!}
	{
        \begin{tabular}{cccc}
            \hline\noalign{\smallskip}
		Variance~($10^{-4}$) & s=2 & s=3 & s=4 \\
            \hline\noalign{\smallskip}
            ResNet18 & \cellcolor{lightgray!30}6.80 & 7.57 & 7.72 \\
            ResNet34 & \cellcolor{lightgray!30}7.25 & 7.95 & 8.10 \\
            \hline
		\end{tabular}
	}
    \end{center}
    \caption{The variance~($10^{-4}$) of the probability distribution output by the pretrained teacher model on the synthetic dataset. The sampling step is fixed to 100. Smaller variances represent smoother probability distributions.}
    \label{table:variance}
    \vspace{-10pt}
\end{table}

\end{document}